\documentclass[twocolumn,natbib]{svjour3}          
\usepackage{color}
\usepackage{tabularx}
\usepackage{graphicx}
\usepackage{amsmath}
\usepackage{amssymb}
\usepackage{enumitem}
\usepackage{xspace}
\usepackage{multirow}
\usepackage{mathtools}
\usepackage{url}
\usepackage{tabulary}
\usepackage[table]{xcolor}
\usepackage{array}
\usepackage[export]{adjustbox}
\usepackage{ctable}
\newcolumntype{Y}{>{\centering\arraybackslash}X}               

\newcommand{\fig}{{Fig.}\@\xspace}
\newcommand{\tab}{{Table}\@\xspace}

\newcommand{\ie}{{i.e.}\@\xspace}
\newcommand{\eg}{{e.g.}\@\xspace}
\newcommand{\etc}{\emph{etc.}}

\journalname{}
\begin{document}
\title
	{
	Visual Social Relationship Recognition
	}
	
\author{
	Junnan~Li\and
	Yongkang~Wong\and
	Qi~Zhao\and
	Mohan~S.~Kankanhalli
}
      
\institute{Junnan~Li \at
	Graduate School for Integrative Sciences and Engineering, National University of Singapore, Singapore, 117456. \\
	\email{lijunnan@u.nus.edu}           
	\and
	Yongkang~Wong \at
	School of Computing, National University of Singapore, Singapore, 117417.
	\email{yongkang.wong@nus.edu.sg}
	\and 
	Qi~Zhao  \at
	Department of Computer Science and Engineering,
	University of Minnesota,
	Minneapolis, MN 55455, USA.\\
	\email{qzhao@cs.umn.edu}  
	\and
	Mohan~S.~Kankanhalli \at 
	School of Computing, National University of Singapore, Singapore, 117417.
	\email{mohan@comp.nus.edu.sg}  
}

\date{}

\maketitle

\begin{abstract}

Social relationships form the basis of social structure of humans.
Developing computational models to understand social relationships from visual data is essential for building intelligent machines that can better interact with humans in a social environment.
In this work, we study the problem of visual social relationship recognition in images.
We propose a Dual-Glance model for social relationship recognition,
where the first glance fixates at the person of interest and the second glance deploys attention mechanism to exploit contextual cues.
To enable this study, 
we curated a large scale People in Social Context (PISC) dataset, 
which comprises of 23,311 images and 79,244 person pairs with annotated social relationships.
Since visually identifying social relationship bears certain degree of uncertainty,
we further propose an Adaptive Focal Loss to leverage the ambiguous annotations for more effective learning.
We conduct extensive experiments to quantitatively and qualitatively demonstrate the efficacy of our proposed method,
which yields state-of-the-art performance on social relationship recognition.

\end{abstract}


\section{Introduction}
\label{sec:introduction}

Since the beginning of early civilizations,
social relationships derived from each individual fundamentally form the basis of social structure in our daily life.
Today,
apart from social interactions that occur in physical world, 
people also communicate through various social media platforms, 
such as Facebook and Instagram.
Large amount of images and videos have been uploaded to the internet that explicitly and implicitly capture people's social relationship information. 
Humans can naturally interpret the social relationships of people in a scene.
In order to build machines with intelligence, 
it is necessary to develop computer vision algorithms that can interpret social relationships.

Enabling computers to understand social relationships from visual data is important for many applications.
First, 
it enables users to pose a socially meaningful query to an image retrieval system, 
such as `Grandma playing with grandson'.
Second, 
visual privacy advisor systems~\citep{Orekondy_ICCV_2017} can alarm users about potential privacy risks if the posted images contain sensitive social relationships.
Third, robots can better interact with people in daily life by inferring people's characteristics and possible behaviors based on their social relationships.
Last but not least, surveillance systems can better analyse human behaviors with the understanding of social relationships.

\begin{figure}[!t]
	\centering
	\begin{minipage}{1.0\columnwidth}
		\centerline{\includegraphics[trim={15 100 15 25},clip,width=1.0\linewidth]{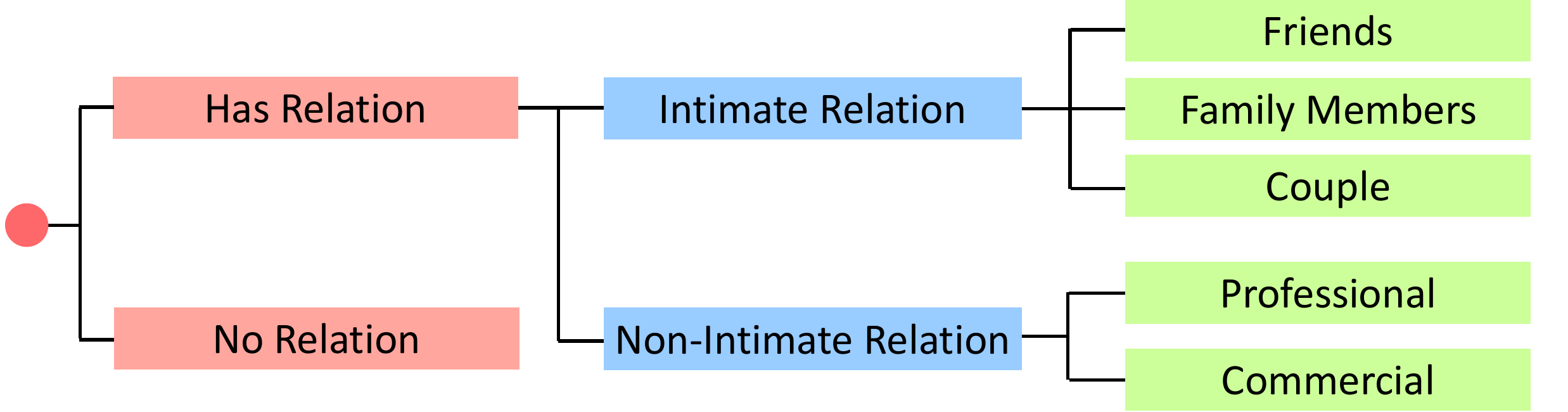}}
	\end{minipage}
  \caption
    {
    \small     
    Defined hierarchical social relationship categories.
    }
  \label{fig:social_relationship_tree}
\end{figure}

In this work, we aim to build computational models that address the problem of visual social relationship recognition in images. 
We start by defining a set of social relationship categories.
With reference to the {\it relational models} theory~\citep{Fiske_PR_1992} in social psychology literature,
we define a hierarchical social relationship categories which embed the coarse-to-fine characteristic of common social relationships
(as illustrated in \fig~\ref{fig:social_relationship_tree}).
Our definition follows a prototype-based approach, 
where we are interested in finding exemplars that parsimoniously describe the most common situations,
rather than an abstract definition that could cover all possible cases.

\begin{figure*}[!t]
	\centering
	\begin{minipage}{\textwidth}
		\centerline{\includegraphics[trim={15 40 15 15},clip,width=1.0\linewidth]{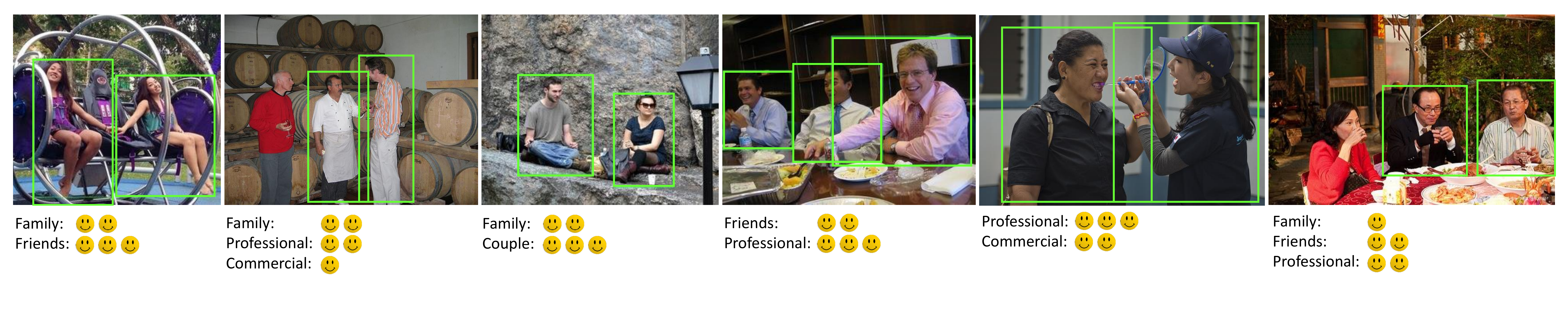}}
	\end{minipage}	
  \vspace{-1ex}
	\caption
		{
		\small  
		Example of images where annotators do not agree on a single social relationship class.
		}
  \label{fig:unsure_example}
\end{figure*}

Social relationship recognition from images is a challenging task for several reasons.
First, images have wide variations in scale, scene, human pose and appearance, as well as occlusions.
Second, humans infer social relationships not only based on the physical appearance (\eg,~color of clothes, gender, age, \etc),
but also from subtler cues (\eg,~expression, proximity, and context) \citep{Alletto_CVPR_2014,Ramanathan_CVPR_2013,Zhang_ICCV_2015}.
Third, a pair of people in an image might have multiple plausible social relationships, as shown in \fig~\ref{fig:unsure_example}.
While previous works on social relationship recognition only consider the majority consensus~\citep{Li_2017_ICCV,Sun_2017_CVPR},
it remains a challenging issue to make use of the ambiguity in social relationship labels.

A preliminary version of this work was published earlier~\citep{Li_2017_ICCV}.
We have extended this work in the following manner:
First, we propose a novel Adaptive Focal Loss, that addresses label ambiguity challenge and class imbalance problem in training.
Second, we improve the Dual-Glance model in~\citep{Li_2017_ICCV} with network modifications (see Section~\ref{sec:attentive-rcnn}).
Third, we conduct additional experiments on two dataset (\ie People in Social Context~\citep{Li_2017_ICCV} and Social Domain and Relation~\citep{Sun_2017_CVPR}),
and achieve significant performance improvement over previous methods. 

The key contributions can be summarized as:
\begin{itemize}
	\item 
	We propose a Dual-Glance model, that mimics the human visual system to explore useful and complementary visual cues for social relationship recognition.
	The first glance fixates at the individual person pair of interest, 
	and performs prediction based on its appearance and geometrical information.
	The second glance exploits contextual cues from regions generated by Region Proposal Network (RPN)~\citep{Ren_NIPS_2015} to refine the prediction.
	\item 
	We propose a novel Attentive R-CNN.
	Given a person pair, 
	the attention is selectively assigned on the informative contextual regions.
	The attention mechanism is guided by both bottom-up and top-down signals.
	%
	\item 
	We propose a novel Adaptive Focal Loss.
	It leverages the embedded ambiguity in social relationship annotations to adaptively modulate the loss and focuses training on hard examples.
	Performance is improved compared to using other loss functions.
	%
	\item 
	To study social relationships,
	we collected the People in Social Context (PISC) dataset.
	It consists of 23,311 images and 79,244 person pairs with manually labeled social relationship labels. 
	In addition, 
	PISC consists of 66 annotated occupation categories.
	\item
	We perform experiments with ablation studies on PISC and the Social Domain and Relation (SDR) \citep{Sun_2017_CVPR} dataset,
	where we quantitatively and qualitatively validate the proposed method.
\end{itemize}

The remainder of the paper is organized as follows.
First, we review the related work in Section~\ref{sec:literature}.
Then we elaborate on the proposed Dual-Glance model in Section~\ref{sec:model}, 
and the Adaptive Focal Loss in Section~\ref{sec:loss}.
Section~\ref{sec:dataset} details the PISC dataset,
whereas the experiment details and results are delineated in Section~\ref{sec:experiment}.
Section~\ref{sec:conclusion} concludes the paper.

\section{Related Work}
\label{sec:literature}

\subsection{Social Relationship}

The study of social relationships lies at the heart of social sciences.
Social relationships are the cognitive sources for generating social action, 
for understanding individual's social behavior, and for coordinating social interaction~\citep{Haslam_JESP_1992}.
There are two forms of representations for relational cognition. 
The first approach represents relationship with a set of theorized or empirically derived dimensions~\citep{Conte_JPSP_1981}.
The other form of representation proposes implicit categories for relation cognition~\citep{Haslam_C_1994}.
One of the most widely accepted categorical theory is the \textit{relational models} theory~\citep{Fiske_PR_1992}.
It offers a unified account of social relations by proposing four elementary prototypes,
namely \textit{communal sharing}, \textit{equality matching}, \textit{authority ranking}, and \textit{market pricing}.
In this work, inspired by the relational models theory, we identify 5 exemplar relationships that are common in daily life and visually distinguishable (\ie~friends, family members, couple, professional and commercial). 
We group them into two relation domains,
namely intimate relation and non-intimate relation, 
as illustrated in \fig~\ref{fig:social_relationship_tree}. 

In the computer vision literature, 
social information has been widely adopted as supplementary cues in several tasks.
\citet{Gallagher_CVPR_2009} extract features describing group structure to aid demographic recognition.
\citet{Shao_ICCV_2013} use social context for occupation recognition in photos.
\citet{Qin_PAMI_2016} exploit social grouping for multi-target tracking.
For group activity recognition,
social roles and relationship information have been implicitly embedded into the inference model~\citep{Choi_ECCV_2012,Deng_CVPR_2016,Direkoglu_ECCV_2012,Lan_CVPR_2012,Lan_PAMI_2012}.
\citet{Alletto_CVPR_2014} define `social pairwise feature' based on F-formation and use it for group detection in egocentric videos. 
Recently, 
\citet{Alahi_2016_CVPR,Robicquet_ECCV_2016} model social factor for human trajectory prediction.

Many studies focus on relationships among family members,
such as siblings, husband-wife, parent-child and grandparent-grandchild.
Such studies include kinship recognition~\citep{Wang_ECCV_2010,Chen_ACMMM_2012,Guo_ICPR_2014,Shao_2014_HCSMA,Xia_TMM_2012} and kinship verification~\citep{Fang_ICIP_2010,Xia_TMM_2012,Hamdi_ICCV_2013} in group photos.
Most of these works leverage facial information to infer kinship, including the location of faces, facial appearance, attributes and landmarks.
\citet{Zhang_ICCV_2015} discover relation traits such as ``warm", ``friendly" and ``dominant" from face images.
Another relevant topic is intimacy prediction~\citep{Yang_CVPR_2012,Chu_ICCV_2015} based on human poses.

For video based social relation analysis,
\citet{Ding_HCSMA_2014} discover social communities formed by actors in movies.
\citet{Manuel_IJCV_2014} detect social interactions in TV shows,
whereas \citet{Yun_CVPR_2012} study human interaction in RGBD videos.
\citet{Ramanathan_CVPR_2013} study social events and discover pre-defined social roles in a weakly supervised setting (\eg birthday child in a birthday party).
\citet{Lv_MMM_2018} propose to use multimodal data for social relation classification in TV shows and movies.
\citet{Fan_CVPR_2018} analyze shared attention in social scene videos.
\citet{Paul_CVPR_2018} construct graphs to understand the relationships and interactions between people in movies.
 
Our study also partially overlaps with the field of social signal processing~\citep{Vinciarelli_TAC_2012},
which aims to understand social signals and social behaviors using multiple sensors.
Such works include interaction detection, role recognition, influence ranking, personality recognition, and dominance detection in group meeting~\citep{GAN_ACMMM_2013,Hung_ACMM_2007,Rienks_ICMI_2006,Salamin_TMM_2009,Alameda_PAMI_2016}.

Very recently, \citet{Li_2017_ICCV,Sun_2017_CVPR} studied social relationship recognition in images.
We \citep{Li_2017_ICCV} propose a Dual-Glance model with Attentive R-CNN to exploit contextual cues,
whereas \citet{Sun_2017_CVPR} leverage semantic attributes learnt from other dataset as intermediate representation to predict social relationships.
Two datasets have been collected, namely the PISC dataset~\citep{Li_2017_ICCV} and the SDR dataset~\citep{Sun_2017_CVPR} (see detailed comparison in Section~\ref{sec:dataset}).
In this paper, we extend our work \citep{Li_2017_ICCV} with Adaptive Focal Loss, improved Dual-Glance model, and additional experiments on both datasets.
 
%

\begin{figure*}[!t]
	\centerline{\includegraphics[width=1.0\textwidth]{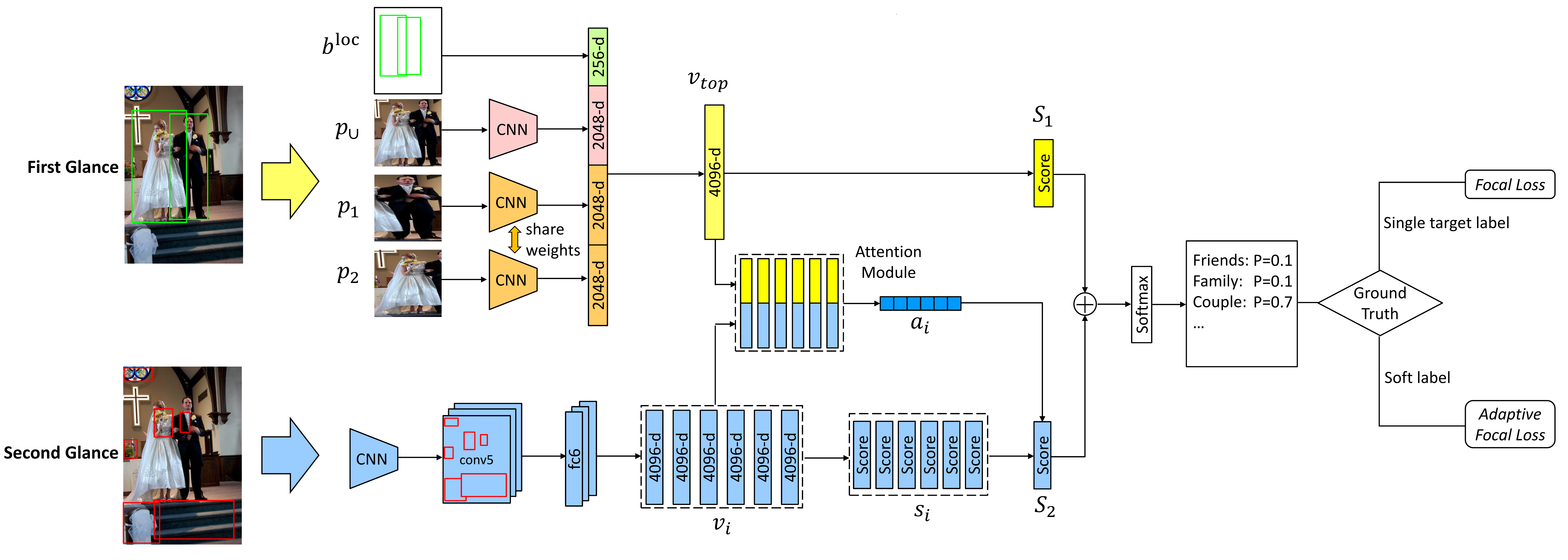}}
	\vspace{-1ex}
  \caption
  	{
  	\small
  	An overview of the proposed Dual-Glance model. 
  	The first glance module fixates at the target person pair and outputs a score. 
  	The second glance module explores contextual regions,
  	allocates attention to each region, 
  	and aggregates regional scores in a weighted manner.
  	The attention is guided by both top-down signal from the first glance, and bottom-up signal form the local region.
  	During training stage, 
  	if the supervision is a hard label (majority vote), we use the focal loss. 
  	If the supervision is a a soft label (distribution over classes), we use the proposed adaptive focal loss. 
  	}
  \label{fig:network}
\end{figure*}

\subsection{Region-based Convolutional Neural Networks}
 
The proposed Attentive R-CNN incorporates Faster R-CNN~\citep{Ren_NIPS_2015} pipeline with attention mechanism to extract information from multiple contextual regions.
The Faster R-CNN pipeline has been widely exploited by many researchers.
\citet{Gkioxari_ICCV_2015} propose R*CNN, that makes use of a secondary region in an image for action recognition.
\citet{Johnson_CVPR_2016} study dense image captioning that focuses on the regions.
\citet{Yikang_2017_ICCV} adopt the Faster R-CNN pipeline as basis framework to study the joint task of object detection, scene graph generation and region captioning.

Attention model has been recently proposed and applied to image captioning~\citep{XU_ICML_2015,You_CVPR_2016}, visual question answering~\citep{Yang_2016_CVPR} 
and fine-grained classification~\citep{Xiao_CVPR_2015}.
In this work, 
we employ attention mechanism on the contextual regions, 
so that each person pair can selectively focus on its informative regions to better exploit contextual cues.
Our attentive R-CNN can also be viewed as a soft Multiple-Instance Learning (MIL) approach~\citep{Maron_NIPS_1997},
where the model receives bags of instances (contextual regions) and bag-level labels (relationship class),
and learns to discover informative instances for correct prediction.

\subsection{Focal Loss}

The proposed Adaptive Focal Loss is inspired by the Focal Loss~\citep{Lin_CORR_2017} for object detection.
Focal loss is designed to address the imbalance in samples between foreground and background classes during training, 
where a modulating factor is introduced to down-weight the easy examples.
Our Adaptive Focal Loss not only addresses class imbalance, 
but more importantly, 
takes into account the uncertainty in visually identifying social relationship labels.

\section{Proposed Dual-Glance Model}
\label{sec:model}

Given an image {$\tens{I}$} and a target person pair highlighted by bounding boxes {$\{b_1,b_2\}$}, 
our goal is to infer their social relationship {$r$}.
In this work, 
we propose a Dual-Glance relationship recognition model,
where the first glance module fixates at $b_1$ and $b_2$, and the second glance module explores contextual cues from multiple region proposals {$\vec{P}_{\tens{I}}$}.
The final score over possible relationships, $\vec{S}$, is computed via
\begin{equation}
\label{eqn:score}
	\vec{S} = \vec{S}_1(\tens{I},b_1,b_2)+\vec{w}\otimes \vec{S}_2(\tens{I},b_1,b_2,\vec{P}_{\tens{I}}),
\end{equation}
where $\vec{w}$ is a weight vector, and {$\otimes$} is the element-wise multiplication of two vectors.
We use softmax to transform the final score into a probability distribution.
Specifically, the probability that a given pair of people having relationship $r$ is calculated as
\begin{equation}
	p_r = \dfrac{\exp(S_r)}{\sum_{r}\exp(S_r)}.
\end{equation}

An overview of the proposed Dual-Glance model is shown in \fig~\ref{fig:network}.

\subsection{First Glance Module}

The first glance module takes in input image {$\tens{I}$} and two human bounding boxes.
First, we crop three patches from {$\tens{I}$} and refer them as {$\tens{p}_{1}$}, {$\tens{p}_{2}$}, and {$\tens{p}_{\cup}$}. 
{$\tens{p}_{1}$} and {$\tens{p}_{2}$} each contains one person, and {$\tens{p}_{\cup}$} contains the union region that tightly covers both people.
The three patches are resized to {$224\times224$} pixels and fed into three CNNs, 
where the CNNs that process {$\tens{p}_{1}$} and {$\tens{p}_{2}$} share the same weights.
The outputs from the last convolutional layer of the CNNs are flattened and concatenated.

We denote the geometry feature of the human bounding box {$b_i$} as
{$\vec{b}_{i}^{\text{loc}} = \{x_{i}^{min},y_{i}^{min},x_{i}^{max},y_{i}^{max},{area}_{i}\} \in \mathbb{R}^5 $},
where all the parameters are relative values, normalized to zero mean and unit variance.
{$\vec{b}_{1}^{\text{loc}}$} and {$\vec{b}_{2}^{\text{loc}}$} are concatenated and processed by a fully-connected (fc) layer. 
We concatenate its output with the CNN features for {$\tens{p}_{1}$}, 
{$\tens{p}_{2}$} and {$\tens{p}_{\cup}$} to form a single feature vector,
which is subsequently passed through another two fc layers to produce first glance score, $\vec{S}_1$.
We use {$\vec{v}_{\text{top}}\in\mathbb{R}^{k}$} to denote the output from the penultimate fc layer.
{$\vec{v}_{\text{top}}$} serves as a top-down signal to guide the attention mechanism in the second glance module.
We set {$k$=4096} with the same dimension as the regional features in Attentive R-CNN.

\subsection{Attentive R-CNN for Second Glance Module}
\label{sec:attentive-rcnn}
For the second glance module, 
we adapt Faster R-CNN \citep{Ren_NIPS_2015} to make use of multiple contextual regions.
Faster R-CNN processes the input image {$\tens{I}$} with Region Proposal Network (RPN) to generate a set of region proposals {$\vec{P}_{\tens{I}}$} with high objectness. 
For each person pair with bounding boxes $b_1$ and $b_2$,
we select the set of contextual regions {$\vec{R}(b_1,b_2;\tens{I})$} from {$\vec{P}_{\tens{I}}$} as
\begin{equation}
\label{eqn:tau}
	\resizebox{0.9\linewidth}{!}{$
	\vec{R}(b_1,b_2;\tens{I}) = \{c \in \vec{P}_{\vec{I}}: \max(G(c,b_1),G(c,b_2)) < \tau_u \}
	$}
\end{equation}

\noindent
where {$G(b_1,b_2)$} computes the Intersection-over-Union (IoU) between two regions,
and {$\tau_u$} is the upper threshold for IoU overlap.
The threshold encourages the second glance module to explore cues different from that of the first glance module.

We then process {$\tens{I}$} with a CNN to generate a convolutional feature map \textit{conv}($\tens{I}$).
For each contextual region {$c\in \vec{R}$}, 
ROI pooling is applied to extract a fixed-length feature vector from \textit{conv}($\tens{I}$),
which is then processed by a fc layer to generate regional feature {$\vec{v}\in\mathbb{R}^{k}$} .
We denote {$\{\vec{v}_i|i=1,2,\ldots,N\}$} as the bag of $N$ regional feature vectors for {$\vec{R}$}.
Each regional feature is then fed to another fc layer to generate a score for the $i$th region proposal:  
\begin{equation}
\vec{s}_i = \tens{W}_s\vec{v}_i+\vec{b}_{s}. 
\end{equation}

Not all contextual regions are informative for the target person pair's relationship.
Therefore we assign different attention to the region scores so that more informative regions could contribute more to the final prediction.
In order to compute the attention,
we first take each local regional feature {$\vec{v}_i$}, and combine it with the top-down feature from the first glance module {$\vec{{v}_{\text{top}}}$} (which contains semantic information of the person pair) into a vector {$\vec{h}_i\in\mathbb{R}^{k}$} via 
\begin{equation}
	\vec{h}_i = \textit{ReLU}(\vec{v}_i+\vec{w_{\text{top}}}\otimes \vec{v_{\text{top}}}), \\	
\end{equation}

\noindent
where {$\vec{{w}_{\text{top}}}\in\mathbb{R}^{k}$}, and {$\otimes$} is the element-wise multiplication.
Then, we calculate the attention $a_i\in [0,1]$ over the $i$th regional score with the sigmoid function:
\begin{equation}
	\begin{split}
	a_i &= \dfrac{1}{1+\exp(-( \tens{W}_{h,a} \vec{h}_i+b_a))},
	\end{split}
\end{equation}
\noindent
where {$\tens{W}_{h,a}\in\mathbb{R}^{1\times k}$} is the weight matrix, and {$b_a\in\mathbb{R}$} is the bias term.

Given the attention, 
the output score of the second glance module is computed as a weighted average of all regional scores:
\begin{equation}
\vec{S}_2 = \dfrac{1}{N}\sum_{i=1}^{N}a_i\vec{s}_i. 
\end{equation}

Note that the Dual-Glance model described above has several differences compared with our previously proposed model~\citep{Li_2017_ICCV}:
(i) We add a new fc6 layer in the Attentive R-CNN model to increase the depth of the network.
(ii) We add \textit{ReLU} non-linearity to compute $\vec{h_i}$, which introduces sparse representation that is more robust. 
(iii) We modify (\ref{eqn:score}) to use element-wise weighting instead of a scalar weight, so that the network can learn to better fuse the scores.
Those modifications can individually improve the performance, and together they lead to +0.7\% improvement in mAP for relationship recognition while other settings remain the same as~\citet{Li_2017_ICCV}.   

%

\section{Adaptive Focal Loss}
\label{sec:loss}

\begin{figure*}[!t]
  \centering
  \begin{minipage}{1\textwidth}
  		\begin{minipage}{0.163\columnwidth} \centerline{\includegraphics[width=\linewidth]{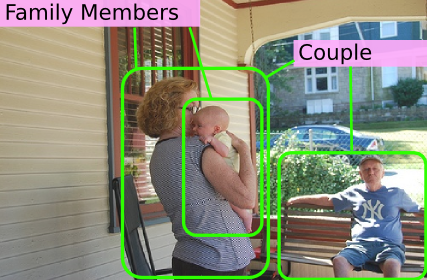}} 			\end{minipage}
  		\begin{minipage}{0.163\columnwidth} \centerline{\includegraphics[width=\linewidth]{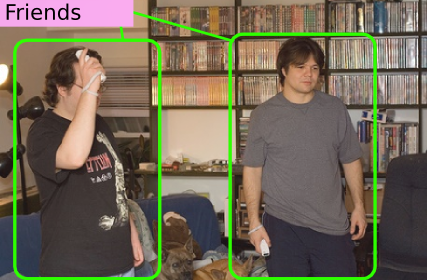}} 			\end{minipage}
  		\begin{minipage}{0.163\columnwidth} \centerline{\includegraphics[width=\linewidth]{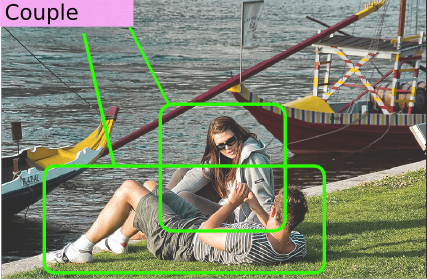}} 			\end{minipage}
  		\begin{minipage}{0.163\columnwidth} \centerline{\includegraphics[width=\linewidth]{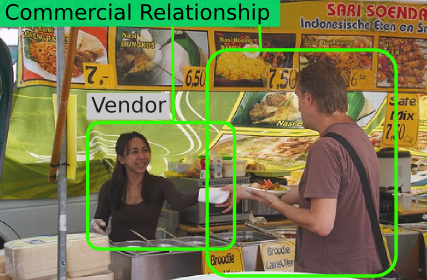}} 	\end{minipage}
  		\begin{minipage}{0.163\columnwidth} \centerline{\includegraphics[width=\linewidth]{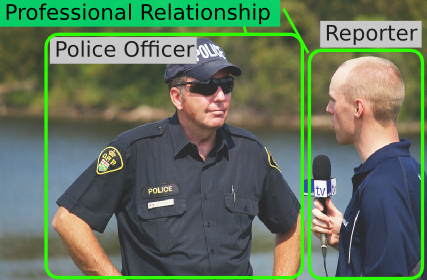}}	\end{minipage}
  		\begin{minipage}{0.163\columnwidth} \centerline{\includegraphics[width=\linewidth]{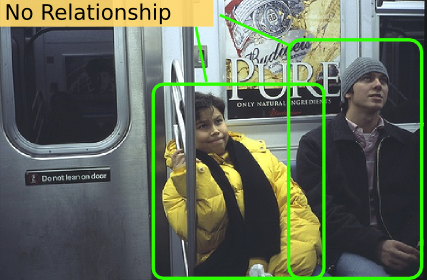}} 					\end{minipage}
  \end{minipage}
  \caption
    {
    \small
		Example images from the People in Social Context (PISC) dataset.
    } 
  \label{fig:image_example}
\end{figure*}

Given a target person pair, our proposed Dual-Glance model outputs a probability distribution $\vec{p}$ over the relationships.
In order to train the model to predict higher probability $p_t$ for the ground truth target relationship $t$,
the standard loss function adopted by~\citet{Li_2017_ICCV,Sun_2017_CVPR} is the cross entropy (CE) loss defined as
\begin{equation}
\label{eqn:CE}
\text{CE}(\vec{p},t) = -\log{p_t}.
\end{equation}

In the task of social relationship recognition,
there often exists class imbalance in the training data.
The classes with more samples can overwhelm the loss and lead to degenerate models.
Previous work addresses this with a heuristic sampling strategy to maintain a manageable balance during training~\citep{Li_2017_ICCV}.
Recently,
in the field of object detection,
focal loss (FL) has been proposed~\citep{Lin_CORR_2017}, 
where a modulating factor $(1-p_t)^\gamma$ is added to the cross entropy loss:
\begin{equation}
\label{eqn:FL}
\text{FL}(\vec{p},t) = -(1-p_t)^\gamma\log{p_t}.
\end{equation}
The modulating factor down-weights the loss contribution from the vast number of well-classified examples, 
and focuses on the fewer hard examples,
where the focusing parameter $\gamma$ adjusts the rate at which easy examples are down-weighted.

In a wide range of visual classification tasks 
(\eg~image classification~\citep{Russakovsky_IJCV_2015}, object detection~\citep{Lin_ECCV_2014}, visual relationship recognition~\citep{Krishna_CoRR_2016}, 
etc.), 
the common approach to determine the ground truth class of a sample is to take the majority vote from human annotations.
While this approach has been effective,
we argue that social relationship recognition is different from other tasks.
The annotation of social relationship has a higher level of uncertainty (as suggested by the agreement rate in Section~\ref{sec:data_stats}), and the minority annotations are not necessarily wrong (as shown in \fig~\ref{fig:unsure_example}).
Therefore, taking the majority vote and ignoring other annotations has the potential disadvantage of neglecting useful information.

In this work, we propose an Adaptive Focal Loss that takes into account the ambiguity in social relationship labels. 
For each sample,
instead of using the hard label from majority voting,
we transform the annotations into a soft label $\vec{p}^y$, which is a distribution calculated by dividing the number of annotations for each relation $r$ with the total number of annotations for that sample.
Then we define the adaptive FL as
\begin{equation}
\label{eqn:Ada-FL}
\text{Ada-FL}(\vec{p},\vec{p}^y) = -\sum_{r}\max((p^y_r-p_r),0)^\gamma\log{p_r}.
\end{equation}

The adaptive FL inherits the ability to down-weight easy examples from the FL,
and extends the FL with two properties to address label ambiguity:
(i)~Instead of considering only the single target class, the adaptive FL takes the sum of losses from all classes, so that all annotations can contribute to training.
(ii)~The modulating factor ${\max((p^y_r-p_r),0)}$ is adaptively adjusted for each class based on the ground truth label distribution. 
The loss still demands the model to predict high probability for the predominant class, but the constraint is relaxed if not all annotations agree.
For example, if 4 out of the 5 annotators agree on \textit{friends} as the label, the adaptive FL term for $\textit{r}=\textit{friends}$ will decrease to 0 if output $p_\textit{friends}\geq 0.8$,
hence it will push $p_\textit{friends}$ to 0.8 instead of 1.
Note that if the ground truth annotations all agree on the same class $t$,
then $p_t^y=1$ and $p_r^y=0$ for $r\neq t$,
the adaptive FL is the same as the FL.

The same philosophy of learning from ambiguous label distributions has also been studied by \citet{Gao_TIP_2017},
where they use the Kullback-Leibler (KL) divergence loss defined as
\begin{equation}
\label{eqn:KL}
\begin{split}
\text{KL-div}(\vec{p},\vec{p}^y) &= \sum_{r}p^y_r\log\dfrac{p^y_r}{p_r}\\
& = -\sum_{r}p^y_r\log{p_r} + \sum_{r}p^y_r\log{p^y_r}\\
& = \text{CE}(\vec{p},\vec{p}^y) - \text{H}(\vec{p}^y),\\
\end{split}
\end{equation}
where {$\text{CE}(\vec{p},\vec{p}^y)$} is the cross entropy between the output distribution and the label distribution,
and {$\text{H}(\vec{p}^y)$} is the entropy of the label distribution.
Since {$\text{H}(\vec{p}^y)$} is independent of the parameters of the model, 
minimizing {$\text{KL-div}(\vec{p},\vec{p}^y)$} is equivalent to minimizing {$\text{CE}(\vec{p},\vec{p}^y)$}.

The difference between KL divergence and the proposed adaptive focal loss is the per-class modulating factor.
While KL divergence uses the ground truth label distribution $p^y_r$ to modulate the per-class loss,
adaptive focal loss uses both $p^y_r$ and the model's output $p_r$ to determine modulation,
thereby down-weighting the easy examples and focusing training on the hard examples.

In practice, similar as~\citet{Lin_CORR_2017}, we use an $\alpha$-balanced variant of the adaptive FL defined as
\begin{equation}
	\resizebox{0.9\hsize}{!}{%
	$\text{Ada-FL}(\vec{p},\vec{p^y}) = -\sum_{r}\alpha_r\max((p^y_r-p_r),0)^\gamma\log{p_r}.$
	}
\end{equation}

$\alpha_r\in [0,1]$ is determined by inverse class frequency via
\begin{equation}
\label{eqn:alpha}
\alpha_r = \left(\dfrac{\min(L_1,L_2,...,L_R)}{L_r}\right)^\beta,
\end{equation}
where $L_r$ is the total number of annotations for relationship $r$, 
and $\beta$ is set to be 0.5 as a smoothing factor.
We find that the $\alpha$-balanced adaptive focal loss yields slightly better performance over the non-$\alpha$-balanced form.

\vspace{-2ex}
\section{People in Social Context Dataset}
\label{sec:dataset}

The People in Social Context (PISC) dataset is an image dataset that focuses on social relationship study (see example images in \fig~\ref{fig:image_example}).
In this section,
we first describe the data curation pipeline.
Then we analyze the dataset statistics and provide comparison with another dataset for social relationship study,
following the presentation style by~\citet{Yash_VQA,Agrawal_VQA}.

\subsection{Curation Pipeline}
\label{sec:collection}

The PISC dataset was curated through a pipeline of three stages. 
In the first stage, 
we collected around 40k images that contain people from a variety of sources, 
including Visual Genome~\citep{Krishna_CoRR_2016}, MSCOCO \citep{Lin_ECCV_2014}, YFCC100M~\citep{Thomee_ACMC_2016},
Flickr, Instagram, Twitter and commercial search engines (\ie Google and Bing).
We used a combination of key words search (\eg~co-worker, people, friends, \etc) and people detector (Faster R-CNN~\citep{Ren_NIPS_2015}) to collect the image.
The collected images have high variation in image resolution, people's appearance, and scene type.

In the second and third stage, 
we hired workers from CrowdFlower platform to perform labor intensive manual annotation task.
The second stage focused on the annotation of person bounding box in each image. 
Following \citet{Krishna_CoRR_2016}, 
each bounding box is required to strictly satisfy the coverage and quality requirements. 
To speed up the annotation process, 
we first deployed Faster R-CNN~\citep{Ren_NIPS_2015} to detect people on all images, 
followed by asking the annotators to re-annotate the bounding boxes if the computer-generated bounding boxes were inaccurately localized.
Overall, 
40\% of the computer-generated boxes are accepted without re-annotation. 
For images collected from MSCOCO and Visual Genome, 
we directly used the provided groundtruth bounding boxes.

Once the bounding boxes of all images had been annotated, 
we selected images consisting of at least two people,
and avoided images that contain crowds of people where individuals cannot be distinguished.
In the final stage,
we requested the annotators to identify the occupation of all individuals in the image,
as well as the social relationships of all person pairs.
To ensure consistency in the occupation categories,
the annotation is based on a list of reference occupation categories.
The annotators could manually add a new occupation category if it was not in the list.

\begin{figure}[!t]
  \centerline{\includegraphics[width=1.0\columnwidth]{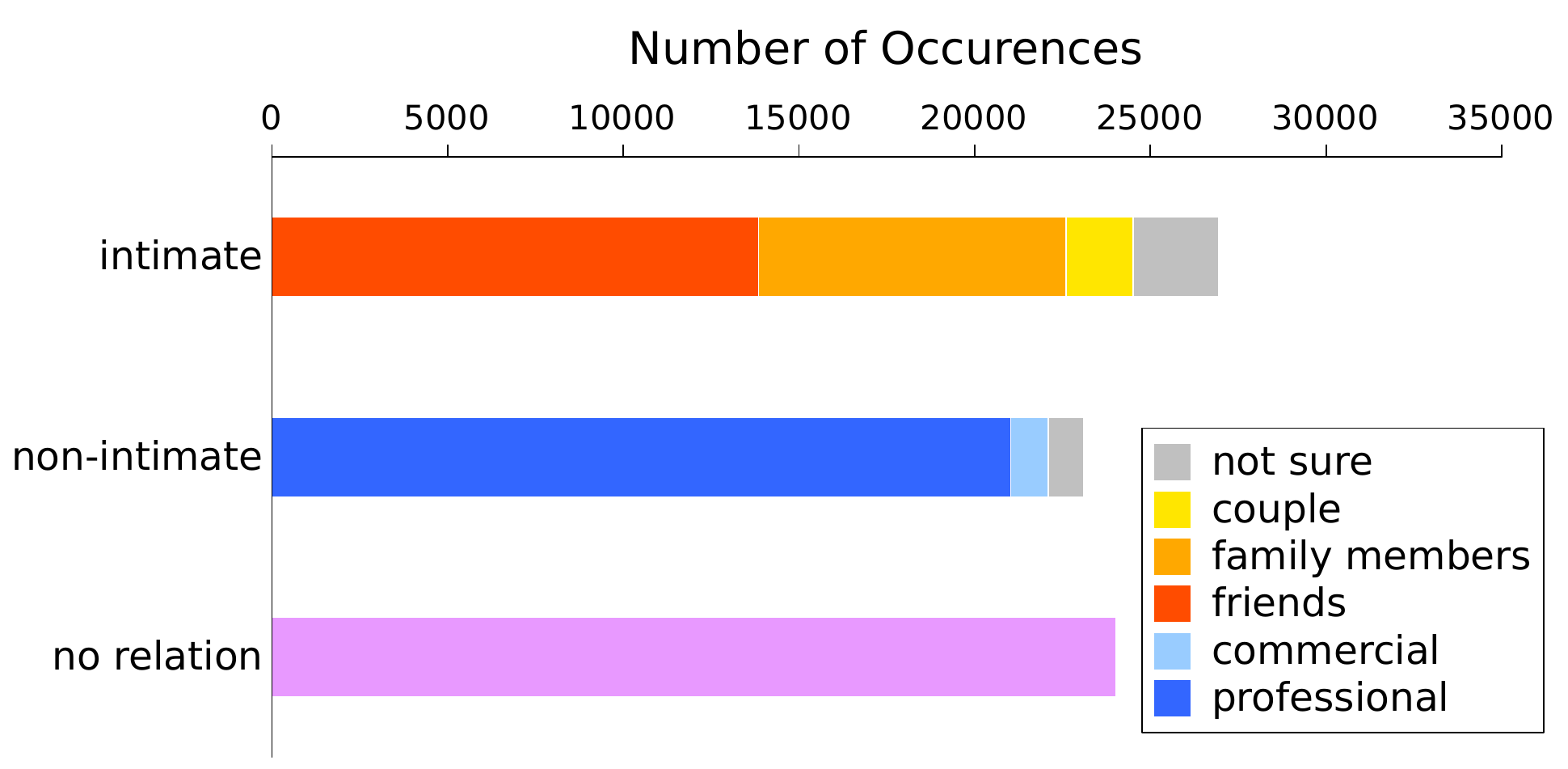}}
  \centerline{\includegraphics[width=1.0\columnwidth]{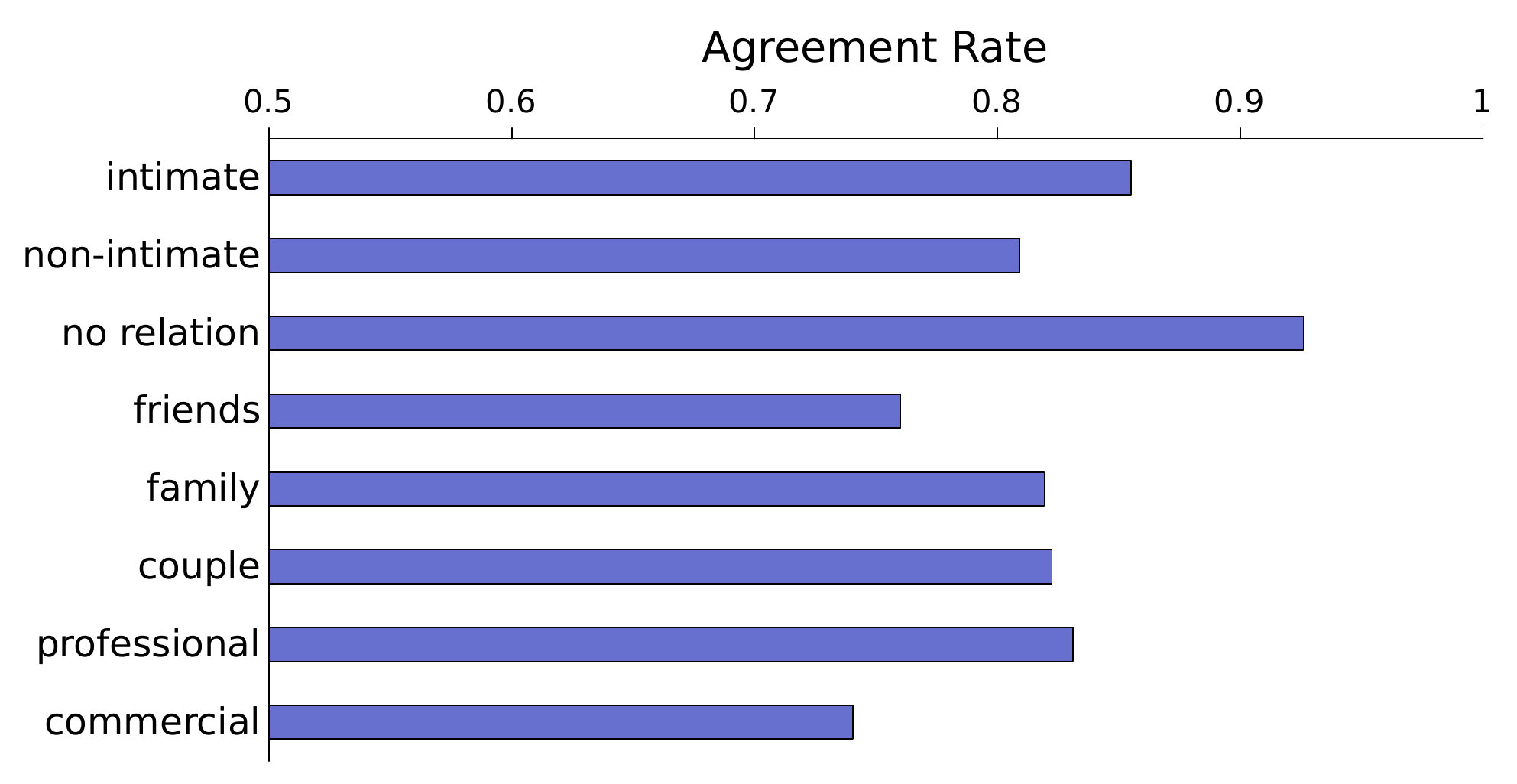}}
	\caption
	  {
	  \small
	  Annotation statistics of the relationship categories.
	  }  
  \label{fig:relation_number}
\end{figure}

\begin{figure}[!t]
  \centerline{\includegraphics[width=0.92\columnwidth]{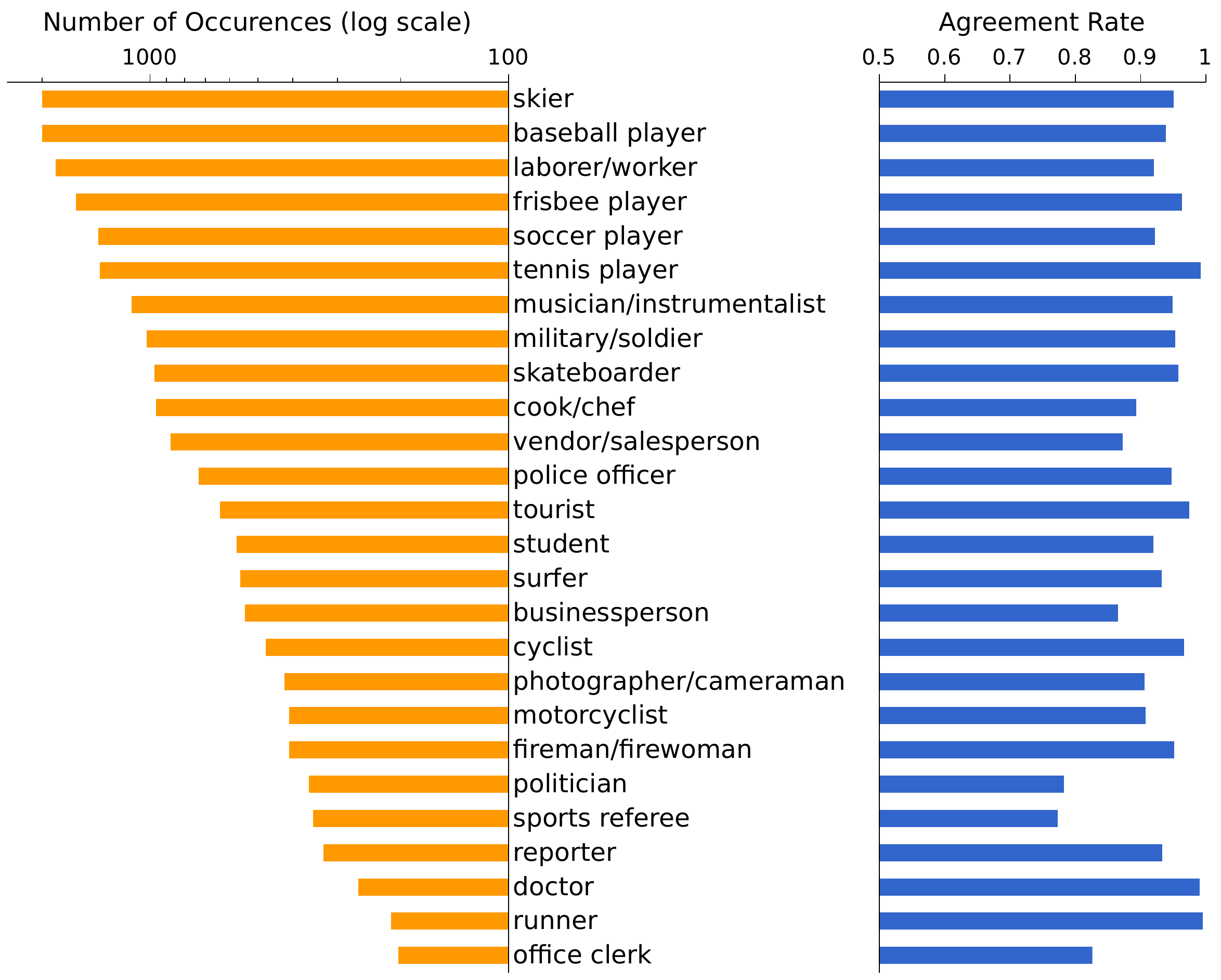}}
  \caption
	  {
	  \small
	  Annotation statistics of the top 26 occupations.
	  }
  \label{fig:occupation_number}
\end{figure}
For social relationships, 
we formulate the annotation task as multi-level multiple choice questions based on the hierarchical structure in \fig~\ref{fig:social_relationship_tree}. 
We provide example images to help annotators understand different relationship classes.
We also provide instructions 
to help annotators distinguish between professional\footnote{
The people are related based on their professions (\eg~co-worker, coach and player, boss and staff, \etc)} and commercial relationship\footnote{
One person is paying money to receive goods/service from the other (\eg~salesman and customer, tour guide and tourist, \etc)}.
Annotators can choose the option `not sure' at any level if they cannot confidently identify the relationship.
Each image was annotated by at least five workers, 
Overall, 
7928 unique workers have contributed to the annotation.

\begin{table*}[!t]
	\centering
	\caption
		{
		\small	
		Comparison between PISC and SDR~\citep{Sun_2017_CVPR} dataset.
		}
	\label{tbl:dataset_compare}
	\vspace{-1ex}
	\begin{tabular}{l|l|l}
		\toprule
		Dataset\hspace{22ex} & PISC & SDR~\citep{Sun_2017_CVPR}\\
		\midrule
		Image source & Wide variety (see Section~\ref{sec:collection}) & Flickr photo album\\
		Number of image & 23,311&8,570\\
		Number of person pair & 79,244 & 26,915\\
		Person's identity & Different images, different people &Multiple images, same person\\
		Person's bounding box& Full-body& Head only\\
		\bottomrule
	\end{tabular}
\end{table*}	

\subsection{Dataset Statistics}
\label{sec:data_stats} 

In total, the PISC dataset consists of 23,311 images with
79,244 pairs of people.
For each person pair,
if there exists a relationship class which at least 60\% of the annotators agree on, 
we refer it as a `consistent' example and assign the majority vote as its class label.
Otherwise we refer it as an `ambiguous' example.
The top part of \fig~\ref{fig:relation_number} shows the distribution of each type of relationships.
We further calculate the agreement rate on the consistent set by dividing the number of agreed human annotations with the total number of annotations.
As shown in the bottom part of \fig~\ref{fig:relation_number},
the agreement rate reflects how visually distinguishable a social relationship class is.
The rate ranges from 74.1\% to 92.6\%,
which indicates that social relationship recognition has certain degree of ambiguity, 
but is a visually solvable problem nonetheless. 




For occupations, 
10,034 images contain people that have recognizable occupations.
In total, 
there are 66 identified occupation categories. 
The occupation occurrence and the agreement rate for the 26 most frequent occupation categories are shown in \fig~\ref{fig:occupation_number}.
Since two source datasets,
\ie~MSCOCO and Visual Genome, 
are highly biased towards `baseball player' and `skier', 
we limit the total number of instances per occupation to 2000 based on agreement rate ranking to ensure there are no bias towards any particular occupation.

\subsection{Comparison with SDR Dataset}

The Social Domain and Relation (SDR) dataset~\citep{Sun_2017_CVPR} is a subset of the PIPA dataset~\citep{Zhang_CVPR_2015} with social relation annotation.
\tab~\ref{tbl:dataset_compare} provides the details of both datasets.
In comparison,
our PISC dataset has multiple advantages.
First and foremost, the PISC dataset contains more images and more person pairs.
Second, the images in SDR dataset all come from Flickr photo albums, while our images are collected from a wide variety of sources.
Therefore, the images in PISC dataset are more diverse.
Third, since the images in SDR dataset were originally collected for the task of people identification~\citep{Zhang_CVPR_2015},
the same person would appear in multiple images, which further reduce the diversity of the data.
Last but not least,
our PISC dataset provides full-body person bounding box annotation,
while SDR dataset provides the head bounding box and uses that to approximate the body bounding box.

\section{Experiment}
\label{sec:experiment}

In this section, we perform experiments and ablation studies to fully demonstrate the efficacy of the proposed method on both PISC and SDR dataset.
We first delineate the dataset and training details,
followed by experiment details and discussion.

\subsection{Dataset Details}
\label{sec:experiment_data}

\noindent\textbf{PISC.} 
On the collected PISC dataset, we perform two tasks,
namely domain recognition (\ie~{\it Intimate} and {\it Non-Intimate}) and relationship recognition (\ie~{\it Friends, Family, Couple, Professional} and {\it Commercial}).
We refer to each person pair as one sample.
For domain recognition, we randomly select 4000 images (15,497 samples) as test set, 4000 images (14,536 samples) as validation set
and use the remaining images (49,017 samples) as training set.
For relationship recognition, since there exists class imbalance in the data,
we sampled the test and validation split to have balanced class.
To do that, 
we select 1250 images (250 per relation) with 3961 samples as test set and
500 images (100 per relation) with 1505 samples as validation set.
The remaining images (55,400 samples) are used as training set.

All the samples used above are selected only from the consistent samples,
where each relationship sample are agreed by a majority of annotators.
For the relationship recognition task,
we enrich the consistent training set with ambiguous samples to create an ambiguous training set.
It contains a total of 58,885 samples, or 3445 samples more than the consistent training set.

\noindent\textbf{SDR.}
The SDR dataset is annotated with 5 domains and 16 relationships~\citep{Sun_2017_CVPR}.
However, the class imbalance is severe for the relationship classes.
7 out of the 16 classes have no more than 40 unique individuals.
In the test set, 4 classes have less than 20 samples (person pairs).
In the validation set, 6 classes have no more than 5 samples.
We tried to re-partition the dataset, but the issue that a same person appears across multiple images makes it very difficult to form a test and validation set with reasonable class balance.
Therefore, we only perform domain recognition task,
where the imbalance is less severe.
The 5 domains include \textit{Attachment, Reciprocity, Mating, Hierarchical power} and \textit{Coalitional groups}.
Note that the samples in SDR dataset are all consistent samples.
\begin{table*}[!t]
	\centering
	\caption
		{
		\small	
		Mean average precision (mAP\%) and per-class recall of baselines and proposed Dual-Glance model on PISC dataset. 
		}
	\label{tbl:fullresult}
	\vspace{-1ex}
	\begin{tabular}{l|c|c|c||c|c|c|c|c|c}	
		\toprule
		 & \multicolumn{3}{c||}{Domain} &	\multicolumn{6}{c}{Relationship}		  \\
		\cmidrule{2-10}
 		Method\hspace{30ex}
 		& \begin{sideways} \parbox{1.8cm}{\raggedright mAP} 	\end{sideways}		
 		& \begin{sideways} \parbox{1.8cm}{\raggedright Intimate} 	\end{sideways} 
 		& \begin{sideways} \parbox{1.8cm}{\raggedright Non-Intimate} 	\end{sideways} 
 		& \begin{sideways} \parbox{1.8cm}{\raggedright mAP} 		\end{sideways}		
 		& \begin{sideways} \parbox{1.8cm}{\raggedright Friends} 	\end{sideways} 
 		& \begin{sideways} \parbox{1.8cm}{\raggedright Family} 	\end{sideways} 
 		& \begin{sideways} \parbox{1.8cm}{\raggedright Couple} 	\end{sideways} 
 		& \begin{sideways} \parbox{1.8cm}{\raggedright Professional} 	\end{sideways} 
 		& \begin{sideways} \parbox{1.8cm}{\raggedright Commercial } 	\end{sideways} 		\\
		\midrule
		Union~\citep{Lu_ECCV_2016}      & 75.2  & 81.5 & 75.3 &49.3 &42.7  & 52.5 &45.0 &70.2 & 49.4 \\
		Location                       & 45.5 & 77.2 & 35.5 &24.1 & 19.4 & 11.9 &46.3 &72.1 & ~~3.5 \\	  	  	  	  
		Pair~\citep{Sun_2017_CVPR}      & 76.9 & 82.1 & 76.5 & 54.9&  58.1 & 58.5 &47.3 &72.7 &52.3  \\	  	
		  	  	  
		Pair+Loc.                      & 77.7  & 82.7 & 77.2 &56.9 & 42.9 & 60.4 &61.7 &80.3 & 54.8 \\		  	    	 
		Pair+Loc.+Union (First-Glance) & \bf{80.2}  &  83.4 &  78.6 & \bf{58.7} &  45.2 & 68.4 & 67.2 &78.3 & 58.8\\
		Pair+Loc.+Global               & 79.4  & 83.1 & 78.5 &58.3 & 44.1 &68.2 &65.4 & 81.1 &57.8\\		  	
		R-CNN                          & 76.0 &81.7 &76.6 & 53.6   &55.7 & 57.8 & 31.6 & 85.5& 42.6 \\	
		All Attributes~\citep{Sun_2017_CVPR} &78.1&82.9&77.6&
		57.5&46.5&59.7&63.2&80.1&55.0\\	
		\midrule
		Dual-Glance                    & 85.4 &85.5 &83.1 & 65.2  & 60.6 & 64.9 &54.7 & 82.2 & 58.0 \\	  
		Dual-Glance + Occupation & \bf{85.8}&85.8&83.5&
		\bf{65.9}&60.1&63.6&55.2&87.9&61.1 \\	
		Dual-Glance + All Attributes & 85.5 &85.2&83.6&
		65.4&58.9&67.8&59.4&81.5&57.7 \\	
		\bottomrule
	\end{tabular}
\end{table*}

\subsection{Training Details}

In the following experiments,
we experiment with both \textit{Focal Loss} using hard label as supervision and the proposed \textit{Adaptive Focal Loss} using soft label distribution as supervision.
We set the focusing parameter $\gamma$ to be 2 in focal loss and 1 in adaptive focal loss, which yield best performance respectively.
Unless otherwise specified,
Section~\ref{sec:experiment_model} uses focal loss on the consistent training set,
Section~\ref{sec:experiment_loss} experiments with various loss functions,
Section~\ref{sec:experiment_tau}-\ref{sec:experiment_att} use adaptive focal loss on the ambiguous training set.

We employ pre-trained CNN models to initialize our Dual-Glance model.
For the first glance, we fine-tune the ResNet-101 model~\citep{He_2016_CVPR}.
For the second glance, we fine-tune the Faster R-CNN model with VGG-16 as backbone~\citep{Ren_NIPS_2015}. 
We employ two-stage training, where we first train the first-glance model until the loss converges,
then we freeze the first-glance model, and train the second-glance model.
We train our model with Stochastic Gradient Descent and backpropagation.
We set learning rate as 0.01, batch size as 32, and momentum as.
During training, we use two data augmentation techniques:
(1) horizontally flipping the image,
and (2) reversing the input order of a person pair 
(\ie~if $p_1$ and $p_2$ are a couple, then $p_2$ and $p_1$ are also a couple.). 

\begin{figure*}[!t]
  \centering
	\begin{minipage}{2.0\columnwidth}
		\centerline{\includegraphics[width=1.0\linewidth]{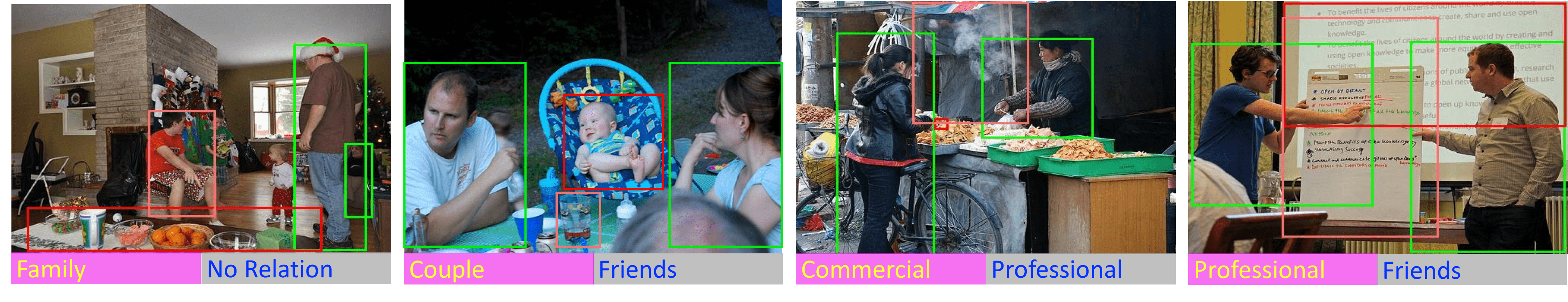}}
	\end{minipage}  
	\vspace{-2ex}
  \caption
    {
    \small
    Examples where Dual-Glance correctly predict the relationship (yellow label) while First-Glance fails (blue label). 
     \textcolor{green}{GREEN} boxes highlight target people pair,
    and the top two contextual regions with highest attention are shown in \textcolor{red}{RED}.
    }
  \label{fig:dual_correct}
\end{figure*}
\begin{figure}[!t]
  \centering
	\begin{minipage}{1.0\columnwidth}
		\centerline{\includegraphics[trim={70 0 50 20},clip,width=0.84\linewidth]{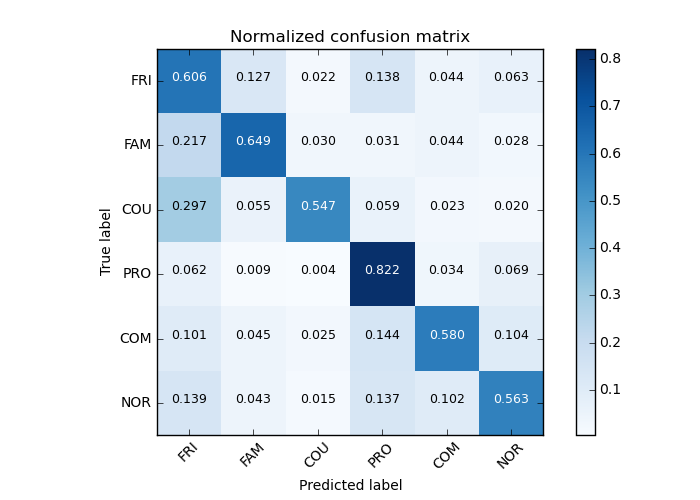}}
	\end{minipage}  
	\vspace{-1ex}
  \caption
    {
    \small
    Confusion matrix of relationship recognition task using the proposed Dual-Glance model trained on PISC dataset.        
    }
  \label{fig:confusion_dual_6class}
\end{figure}

\subsection{Baselines vs. Dual-Glance}
\label{sec:experiment_model}

We evaluate multiple baselines and compare them to the proposed Dual-Glance model to show its efficacy.
Formally, 
the compared methods are as followed:
\begin{enumerate}
	\item 
	\textbf{Union}: 
	Following the predicate prediction model by~\citet{Lu_ECCV_2016},
	we use a CNN model that takes the union region of the person pair as input, and outputs their relationship.
	%
	%
	\item
	\textbf{Location}:
	We only use the geometry feature of the two individuals' bounding boxes to infer their relationship.
	\item
	\textbf{Pair}: 
	The model consists of two CNNs with shared weights. The inputs are two cropped image patches for the two individuals.
	The model is similar to the \textbf{End-to-end Finetuned} double-stream CaffeNet in~\citep{Sun_2017_CVPR}, except that~\citet{Sun_2017_CVPR} don't share weights.
	\item
	\textbf{Pair+Loc.}:
	We extend Pair by using the geometry feature of the two bounding boxes.
	\item
	\textbf{Pair+Loc.+Union}: 
	First-Glance model illustrated in \fig~\ref{fig:network}, which combines Pair+Loc. with Union.
	\item
	\textbf{Pair+Loc.+Global}:
	Model structure is the same as first-glance, except that we replace the union region with the entire image as global input.  
	\item
	\textbf{R-CNN}:
	We train a R-CNN using the region proposals {$\vec{R}(b_1,b_2;\tens{I})$} in (\ref{eqn:tau}), and use average pooling to combine the regional scores.	
	\item
	\textbf{All Attributes}~\citep{Sun_2017_CVPR}:
	We follow the method by~\cite{Sun_2017_CVPR} and extract 9 semantic attributes (age, gender, location\&scale, head appearance, head pose, face emotion, clothing, proximity, activity) using models pre-trained on multiple annotated datasets. 
	Then a linear SVM is used for classification.
	The SVM is calibrated to produce probabilities for calculating mAP.
	For attributes that require head bounding boxes (\eg age, head pose, face emotion, etc.), we use a pre-trained head detector to find the head bounding box within each person's ground-truth body bounding box.

	\item
	\textbf{Dual-Glance}:
	Our proposed model (Section~\ref{sec:model}). 
	\item
	\textbf{Dual-Glance+Occupation}:
	We first train a CNN for occupation recognition using the collected occupation labels.
	Then during social relationship training,
	we concatenate the occupation score (from the last layer of the trained CNN) for each person with the human-centric feature $\vec{v}_{\text{top}}$ as the new human-centric feature for the first glance.	
	\item
	\textbf{Dual-Glance+All Attributes}:
	We fuse the score from baseline 8 with the score from the dual-glance model for the final prediction.
\end{enumerate}

\tab~\ref{tbl:fullresult} shows the results for both domain recognition task and relationship recognition task on the PISC dataset.
We can make several observations from the results.
First, {\small\textbf{Pair+Loc.}} outperforms {\small\textbf{Pair}}, which suggests that peoples' geometric location in an image contains information useful to infer their social relationship.
This is supported by the law of \textit{proxemics}~\citep{Hall_1959_silent} which says people's interpersonal distance reflects their relationship.
However, the location information alone cannot be used to predict relationship, as shown by the results of {\small\textbf{Location}}.
Second,
adding {\small\textbf{Union}} to {\small\textbf{Pair+Loc.}} improves performance. 
The performance gain is lesser if we use the global context (entire image) rather than the union region.
Third, using contextual regions is effective for relationship recognition.
{\small\textbf{R-CNN}} achieves comparable performance to the first-glance model by using only contextual regions.
The proposed {\small\textbf{Dual-Glance}} model outperforms the first-glance model by a significant margin ($+5.2\%$ for domain recognition, $+6.5\%$ for relationship recognition).

Visual attributes also provide useful mid-level information for social relationship recognition. 
Combining {\small\textbf{All Attributes}} with {\small\textbf{Dual-Glance}} slightly improves performance,
while {\small\textbf{Dual-Glance+Occupation}} achieves the best performance among all methods.
However, {\small\textbf{All Attributes}} itself cannot outperform the proposed first-glance method.
The reason is because of the unreliable attribute detection caused by frequently occluded head/face in the PISC dataset or the domain shift from source datasets (where the attribute detectors are trained) to target dataset (where the attribute detectors are applied, \ie~PISC).
\begin{table}[!t]
	\centering
	\caption
		{
		\small	
		Domain recognition result (\%) on SDR dataset.
		}
	\label{tbl:PIPA}
	\vspace{-1ex}
	\begin{tabular}{l|c} 
		\toprule
		Method\hspace{15ex}                 & Accuracy		\\
		\midrule
		End-to-end Finetuned~\citep{Sun_2017_CVPR} & 59.0\\ 
		All Attributes~\citep{Sun_2017_CVPR} & 67.8 \\
		First-Glance                        & 68.2 \\
		Dual-Glance                         & 72.1 \\
		Dual-Glance+All Attributes&  \bf{72.5} \\
		\bottomrule
	\end{tabular}
\end{table}	

\begin{table*}[!t]
	\centering
	\caption
		{
		\small	
		Relationship recognition result (mAP\%) on PISC dataset with various loss functions and different level of ambiguity in training data.
		* is the optimal results in Table~\ref{tbl:fullresult}.
		}
	\label{tbl:adaptive_focal}
	\vspace{-1ex}
	\begin{tabular}{l|l|l|c|c} 
		\toprule
		Loss Function\hspace{10ex} & Training Set\hspace{3ex} & Training Supervision\hspace{3ex} & ~~First-Glance~~ & ~Dual-Glance~ 		\\
		\midrule
		Cross Entropy       & \multirow{4}{*}{Consistent} & Single label $t$               & 57.4&63.9\\
		Focal Loss*          &                             & Single label $t$               & 58.7  & 65.2  \\
		KL divergence       &                             & Soft label $\Vec{p^y}$ & 58.7 & 65.1 \\		
		Adaptive Focal Loss &                             & Soft label $\Vec{p^y}$ & \bf{59.7} &  \bf{66.4}   \\		
		\midrule
		KL divergence       & \multirow{2}{*}{Ambiguous}  & Soft label $\Vec{p^y}$ & 59.1 & 65.8 \\		
		Adaptive Focal Loss &                             & Soft label $\Vec{p^y}$ & \bf{61.2} & \bf{68.3} \\	
		\bottomrule
	\end{tabular}
 	\vspace{-1ex}
\end{table*}	
\fig~\ref{fig:dual_correct} shows some intuitive illustrations where the dual-glance model correctly classifies relationships that are misclassified by the first-glance model.

\fig~\ref{fig:confusion_dual_6class} shows the confusion matrix of relationship recognition with the proposed Dual-Glance model, where we include \textit{no relation} (NOR) as the 6th class.
The model tends to confuse the intimate relationships, 
especially, misclassifying \textit{family} and \textit{couple} as \textit{friends}.

Table~\ref{tbl:PIPA} shows the result of domain recognition task on SDR dataset.
{\small\textbf{End-to-end Finetuned}}~\citep{Sun_2017_CVPR} is a double-stream CNN model that uses the person pair as input, similar to our {\small\textbf{Pair}} except for weight sharing. 
{\small\textbf{All Attributes}} is the best-performing method by~\citet{Sun_2017_CVPR},
where a set of pretrained models from other dataset are used to extract semantic attribute representations (\eg age, gender, activity, etc.),
and a linear SVM is trained to classify relation using the semantic attributes as input.
Compared with the results from~\citet{Sun_2017_CVPR}, 
both our First-Glance and Dual-Glance yield better performance.
While First-Glance slightly outperforms {\small\textbf{All Attributes}},
Dual-Glance achieves more improvement by utilizing contextual regions.


\subsection{Efficacy of Adaptive Focal Loss}
\label{sec:experiment_loss}

In this section,
we conduct relationship recognition experiment on the PISC dataset using various loss functions and two training data.
We experiment with cross entropy loss (\ref{eqn:CE}),
focal loss (\ref{eqn:FL}),
KL divergence loss (\ref{eqn:KL}) and the proposed adaptive focal loss (\ref{eqn:Ada-FL}) on both consistent training set and ambiguous training set (see Section \ref{sec:experiment_data} for dataset details).
Note that we use the $\alpha$-balanced version for all losses while $\alpha$ is computed as in (\ref{eqn:alpha}).

Table~\ref{tbl:adaptive_focal} shows the result.
There are several observations we can make.
First, comparing cross entropy loss and focal loss that both use single target label as training supervision,
focal loss yields better performance ($+1.3\%$).
Second,
adaptive focal loss achieves further improvement on focal loss.
With dual-glance model, the improvement is $+1.2\%$ in mAP if we use the same consistent training set.
If we train on the ambiguous set, the improvement boosts to $+3.1\%$.
Third,
KL-divergence loss produces similar performance compared to focal loss on consistent set,
and slight improvement on ambiguous set.
On both training sets, KL-divergence gives lower mAP compared to adaptive focal loss.
And last but not least, compared with the cross entropy loss by~\citet{Li_2017_ICCV},
the proposed adaptive focal loss with ambiguous training set increases mAP by \textbf{+4.4\%} using Dual-Glance model.
The results demonstrate that the minority social relationship annotations do contain useful information,
and the proposed adaptive focal loss can effectively exploit the ambiguous annotations for more accurate relationship recognition.

\begin{figure}[!t]
  \centering
	\begin{minipage}{1.0\columnwidth}
		\begin{minipage}{0.49\columnwidth} \centerline{\includegraphics[width=1.0\linewidth]{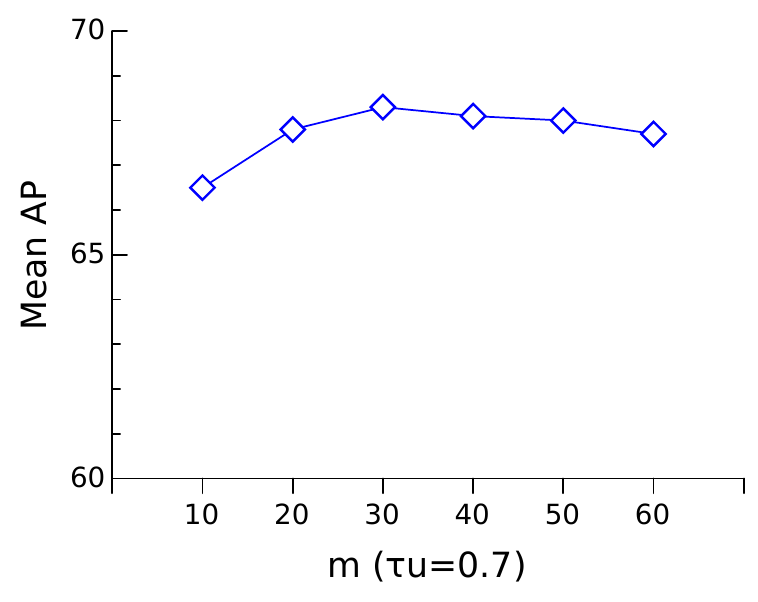}} \end{minipage}
		\begin{minipage}{0.49\columnwidth} \centerline{\includegraphics[width=1.0\linewidth]{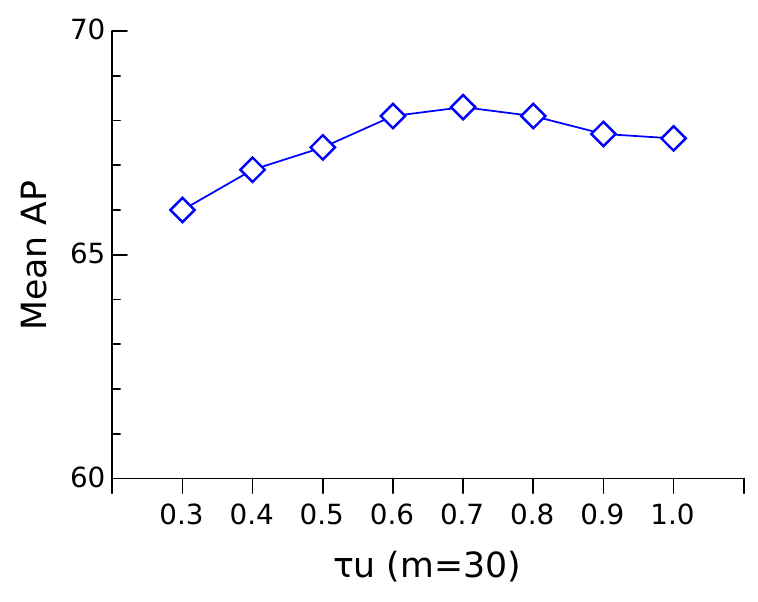}} \end{minipage}
	\end{minipage}
	\vspace{-1ex}
  \caption
    {
    \small
		Performance of Dual-Glance model on PISC dataset over variations in maximum number of region proposals (Left) and upper threshold of overlap between proposals and the person pair (Right).
    }
  \label{fig:regions}  
\end{figure}
\subsection{Variations in Contextual Regions}
\label{sec:experiment_tau}

In order to encourage the attentive R-CNN to explore contextual cues that are not used by First-Glance,
we set a threshold $\tau_u$ in (\ref{eqn:tau}) to suppress regions that highly overlap with the person pair.
Another influence factor in attentive R-CNN is the number of region proposals $m$ from RPN, 
which can be controlled by a threshold on the objectness score.
In this section, 
We experiment with different combinations of $m$ and $\tau_u$ with the dual-glance model trained using adaptive focal loss on PISC dataset.
As shown in \fig~\ref{fig:regions}, $m=30$ and $\tau_u=0.7$ produce the best performance on relationship recognition.

\begin{table}[!t]
	\centering
	\caption
		{
		\small	
		Relationship recognition result (mAP\%) using different person bounding box on PISC dataset.
		}
	\label{tbl:automatic_det}
	\begin{tabular}{l|c|c} 
		\toprule
		Method\hspace{8ex} & Ground Truth&Faster R-CNN \\
	  \midrule
		First-Glance &	61.2& 59.7\\	
		Dual-Glance & 68.3 & 67.5\\
	  \bottomrule
	\end{tabular}
\end{table}	

\subsection{Ground Truth vs. Automatic People Detection}
\label{sec:experiment_box}

In this section, we study the propose method using ground truth annotation of person's bounding box. 
In other words we assume to possess a person detector that works as well as human.
In this section, we test the robustness of our proposed method with automatic person detector.
We employ Faster R-CNN~\citep{Ren_NIPS_2015} person detector pre-trained on MSCOCO dataset.
Same as \citet{Ren_NIPS_2015}, for each person in the test set,
we treat all output boxes with $\nless 0.5$ IoU overlap with the ground truth box as positives,
and apply greedy non-maximum suppression to select the highest scoring box as final prediction.
In total,
3171 out of 3961 person pairs have been detected, while the average IoU overlap between detection boxes and ground truth is 79.7\%.

Table \ref{tbl:automatic_det} shows the relationship recognition result.
Using automatic person detector leads to $-1.5\%$ decrease in mAP for first-glance model.
The decrease is slighter for dual-glance model ($-0.8\%$), because the attentive R-CNN is less affected by person's bounding box.
The relatively insignificant performance decrease indicates that our proposed model is robust to person detection noise,
and can be applied in a fully automatic setting.

\subsection{Analysis on Attention Mechanism}
\label{sec:experiment_att}

In this section we demonstrate the importance of the attention mechanism on the proposed Dual-Glance model.
We remove the attention module and experiment with two functions to aggregate regional scores,
which are $\text{avg}(\cdot)$ and $\max(\cdot)$.
\tab~\ref{tbl:MIL} shows the relationship recognition result on PISC dataset.
Adding attention mechanism leads to improvement for both $\text{avg}(\cdot)$ and $\max(\cdot)$.
The performance improvement is more significant for $\text{avg}(\cdot)$.
For dual-glance without attention, $\max(\cdot)$ performs best,
While for dual-glance with attention, $\text{avg}(\cdot)$ performs best. 
This is because $\max(\cdot)$ assumes that there exists a single contextual region that is most informative of the relationship,
but sometimes there is no such region.
On the other hand,
$\text{avg}(\cdot)$ consider all regions,
but could be distracted by irrelevant ones.
However, with properly guided attention, 
$\text{avg}(\cdot)$ can better exploit the collaborative power of relevant regions for more accurate inference.

\begin{table}[!t]
	\centering
	\caption
		{
		\small	
		Relationship recognition result (mAP\%) of the proposed Dual-Glance model with and without attention mechanism using various aggregation functions on PISC dataset.
		}
	\label{tbl:MIL}
	\begin{tabularx}{0.9\columnwidth}{Y|Y||Y|Y} 
		\toprule
		 \multicolumn{2}{c||}{Without Attention} &	\multicolumn{2}{c}{With Attention}
		\\
		 {\scriptsize $\text{avg}(\cdot)$}  &{\scriptsize $\max(\cdot)$} &
		 {\scriptsize $\text{avg}(\cdot)$}   &{\scriptsize $\max(\cdot)$}\\
	  \midrule
	 64.0 & 65.5 &\bf{68.3} & 67.1   \\	
	  \bottomrule
	\end{tabularx}
\end{table}	

\begin{figure}[!t]
	\centering
	\begin{minipage}{1.0\columnwidth}
		\centerline{\includegraphics[width=\columnwidth]{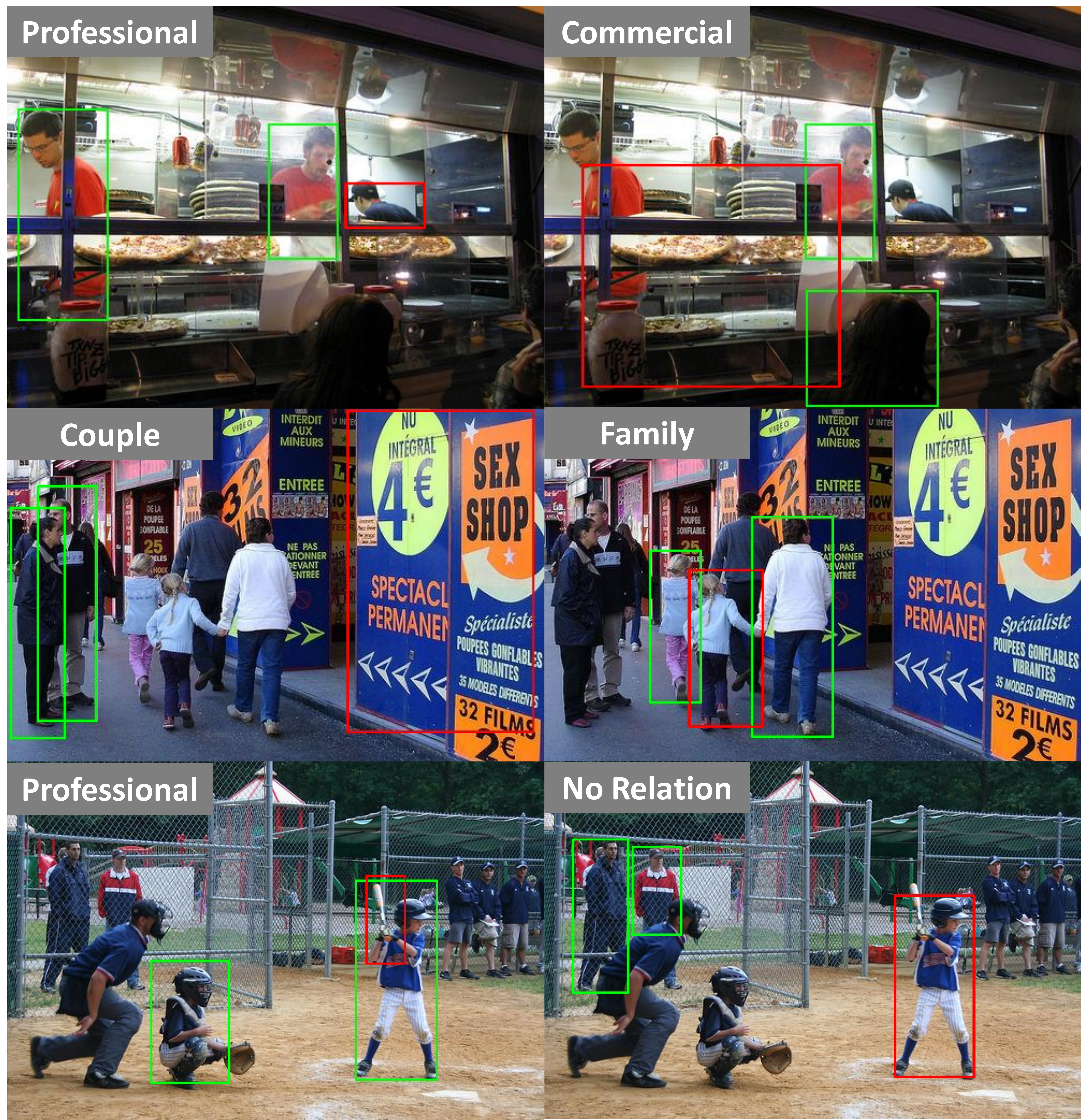}}
	\end{minipage}
  \caption
    {
    \small
    Illustration of the proposed attentive RCNN.
    \textcolor{green}{GREEN} boxes highlight the target pair of people, and \textcolor{red}{RED} box highlights the contextual region with the highest attention.
    For each target pair, the attention mechanism fixates on different region.  
    }
  \label{fig:visualize_attention}
\end{figure}

\begin{figure}[!t]
	\centering
	\begin{minipage}{1.0\columnwidth}
		\centerline{\includegraphics[trim={0 20 0 0},clip,width=1.0\linewidth]{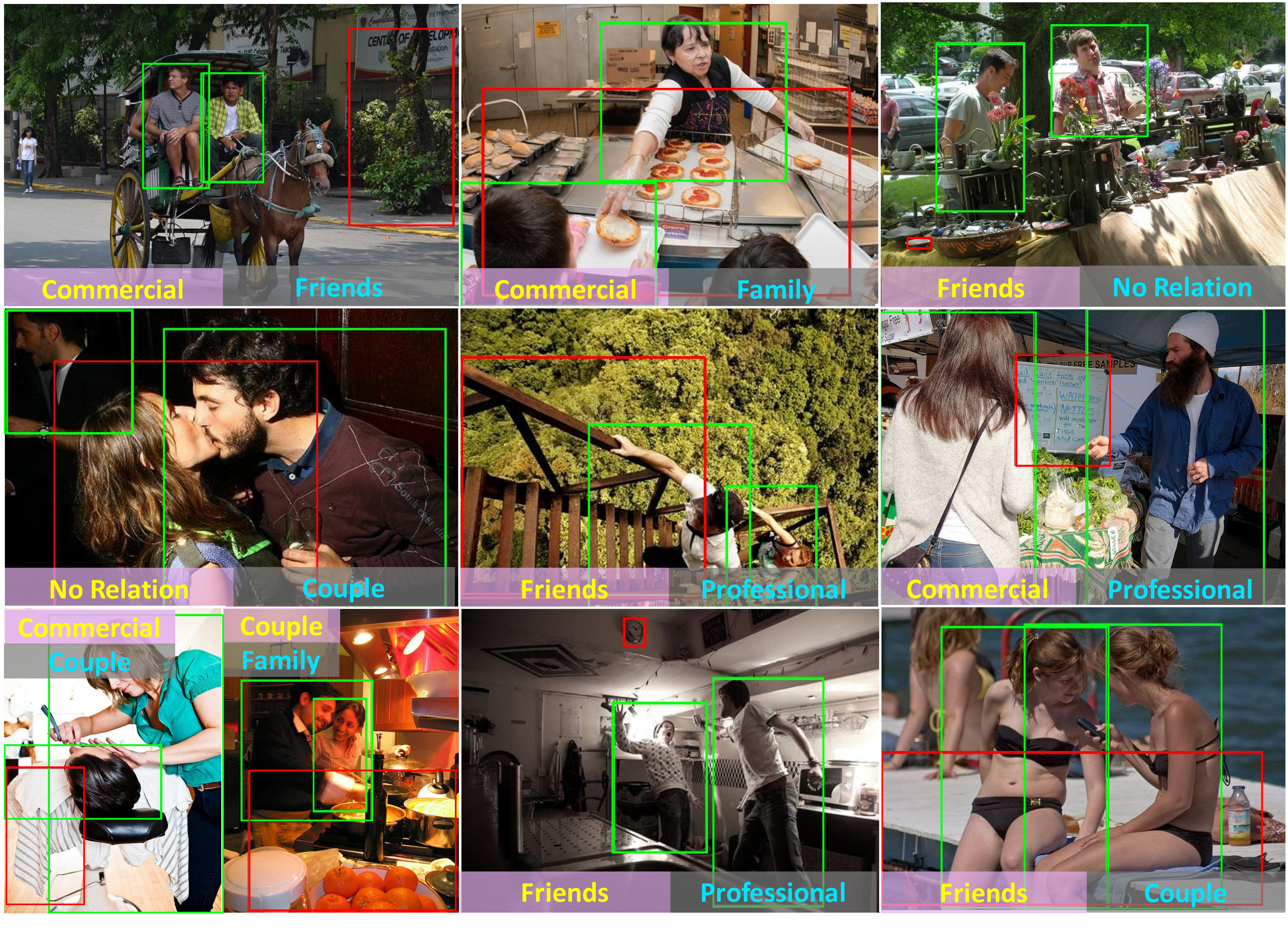}}
	\end{minipage}  
	\vspace{-1ex}
  \caption
    {
    \small
    Examples of incorrect predictions on PISC dataset. 
    Yellow labels are the ground truth, and \textcolor{blue}{BLUE} labels are the model's predictions.
    }
  \label{fig:visualize_wrong}
\end{figure}

\begin{figure*}[!t]
	\centerline{\includegraphics[trim={60 75 60 0},clip,width=1.0\textwidth]{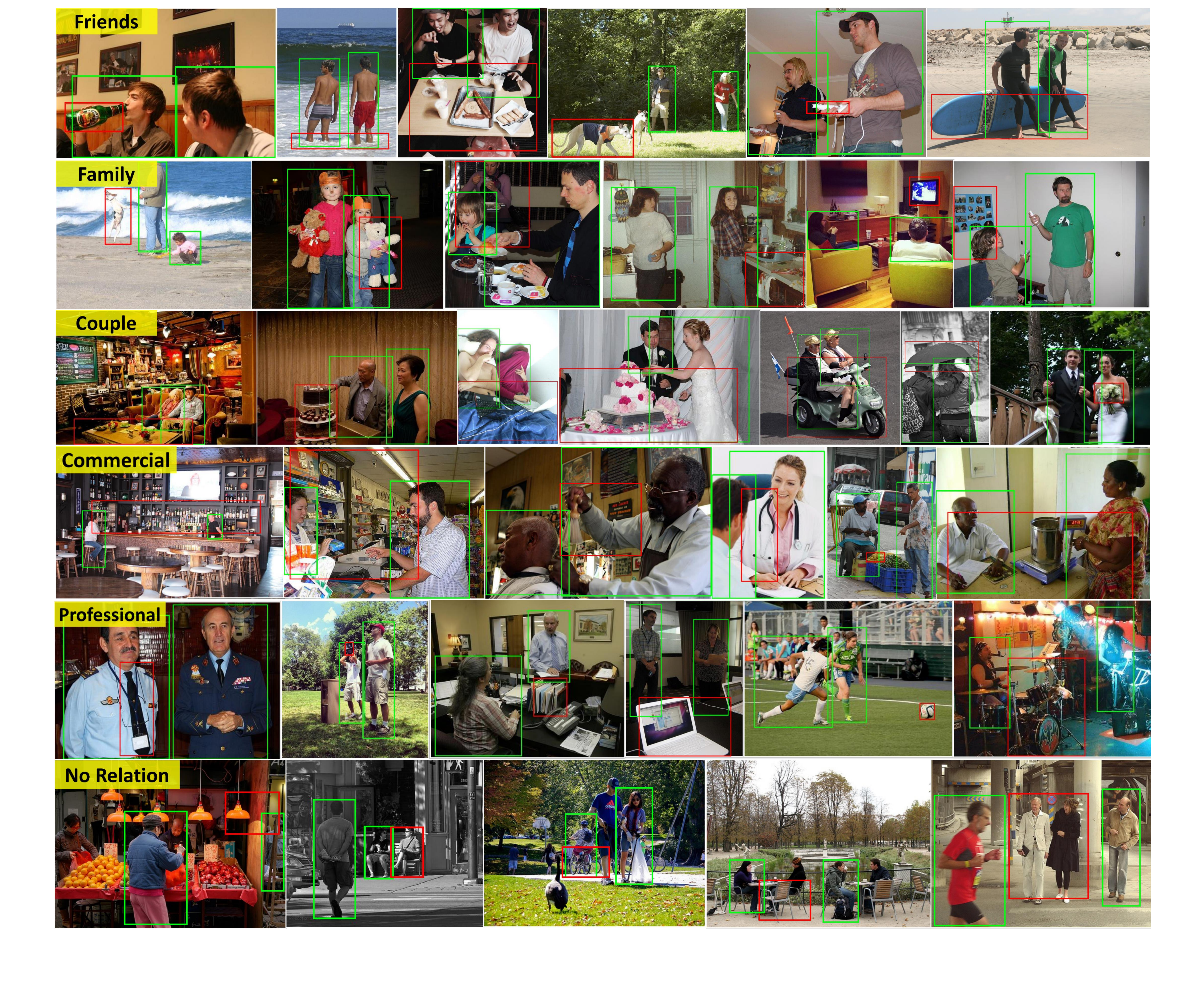}}
  \caption
    {
    \small
    Example of correct predictions on PISC dataset. 
    \textcolor{green}{GREEN} boxes highlight the targets, and \textcolor{red}{RED} box highlights the contextual region with highest attention.
    }
  \label{fig:visualize_correct}
\end{figure*}

\subsection{Visualization of Examples}
\label{sec:visualization}

The attention mechanism enables different person pairs to exploit different contextual cues.
Some examples are shown in \fig~\ref{fig:visualize_attention}.
Taking the images on the second row as an example, the little girl in red box is useful to infer that the other girl on her left and the woman on her right are family,
but her existence indicates little of the couple in black.

\fig~\ref{fig:visualize_wrong} shows examples of the misclassified cases. 
The model fails to pick up gender cue (misclassifies \textit{friends} as \textit{couple} in the image at row 3 column 3),
or picks up the wrong cue (the white board instead of the vegetable in the image at row 2 column 3).
\fig~\ref{fig:visualize_correct} shows examples of correct recognition for each relationship category in the PISC test set. 
We can observe that the proposed model learns to recognize social relationship from a wide range of visual cues including clothing, environment, surrounding people/animals, contextual objects, etc.
For intimate relationships, the contextual cues varies from \textit{beer} (friends),
\textit{gamepad} (friends),
\textit{TV} (family), to \textit{cake} (couple) and \textit{flowers} (couple).
In terms of non-intimate relationships, the contextual cues are mostly related to the occupations of the individuals.
For instance, \textit{goods shelf} and \textit{scale}  indicate commercial relationship, 
while \textit{uniform} and \textit{documents} imply professional relationship.

\section{Conclusion}
\label{sec:conclusion}

In this study, we address the problem of social relationship recognition, 
a key challenge to bridge the social gap towards higher-level social scene understanding.
To this end, we propose a dual-glance model, which exploits useful information from the person pair of interest as well as multiple contextual regions.
We incorporate attention mechanism to assess the relevance of each region instance with respect to the person pair.
We also propose an adaptive focal loss, that leverages the ambiguity in social relationship labels for more effective learning.
The adaptive focal loss can be potentially used in a wider range of tasks that have a certain degree of subjectivity, such as sentiment classification, aesthetic prediction, image style recognition, etc.

In order to facilitate research in social scene understanding,
we curated a large-scale PISC dataset.
We conduct extensive experiments and ablation studies,
and demonstrate both quantitatively and qualitatively the efficacy of the proposed method.
Our code and data are available at \url{https://doi.org/10.5281/zenodo.831940}.

Our work builds a state-of-the-art computational model for social relationship recognition.
We believe that our work can pave the way to more studies on social relationship understanding, and social scene understanding in general.

 
\section*{Acknowledgment}
\label{sec:acknowledgement}

This research was carried out at the NUS-ZJU SeSaMe Centre.
It  is supported by the National Research Foundation, 
Prime Minister's Office, 
Singapore under its International Research Centre in Singapore Funding Initiative. 

\bibliographystyle{spbasic}
\bibliography{bibliographie}

\end{document}